\documentclass[letterpaper]{article} % DO NOT CHANGE THIS
\usepackage[submission]{aaai25}  % DO NOT CHANGE THIS
\usepackage{times}  % DO NOT CHANGE THIS
\usepackage{helvet}  % DO NOT CHANGE THIS
\usepackage{courier}  % DO NOT CHANGE THIS
\usepackage[hyphens]{url}  % DO NOT CHANGE THIS
\usepackage{graphicx} % DO NOT CHANGE THIS
\usepackage{natbib}  % DO NOT CHANGE THIS AND DO NOT ADD ANY OPTIONS TO IT
\usepackage{caption} % DO NOT CHANGE THIS AND DO NOT ADD ANY OPTIONS TO IT
\usepackage{algorithm}
\usepackage{algorithmic}
\usepackage{amsmath}
\usepackage{algorithm}
\usepackage{algorithmic}

\usepackage{mathtools}
\usepackage{amsthm}
\usepackage{bm}

\usepackage[utf8]{inputenc} % allow utf-8 input
\usepackage[T1]{fontenc}    % use 8-bit T1 fonts
\usepackage{url}            % simple URL typesetting
\usepackage{booktabs}       % professional-quality tables
\usepackage{amsfonts}       % blackboard math symbols
\usepackage{nicefrac}       % compact symbols for 1/2, etc.
\usepackage{microtype}      % microtypography
\usepackage{xcolor}    
\usepackage{newfloat}
\usepackage{listings}
\DeclareCaptionStyle{ruled}{labelfont=normalfont,labelsep=colon,strut=off} % DO NOT CHANGE THIS
\floatstyle{ruled}
\newfloat{listing}{tb}{lst}{}
\floatname{listing}{Listing}
\title{Flexible Bayesian Last Layer Models Using Implicit Priors and
Diffusion Posterior Sampling}
\author{
    Jian Xu\textsuperscript{\rm 1},
    Zhiqi Lin\textsuperscript{\rm 1},
    Shigui Li\textsuperscript{\rm 1},
    Min Chen\textsuperscript{\rm 1},
    Junmei Yang\textsuperscript{\rm 1},\\
    Delu Zeng\textsuperscript{\rm 1},
    John Paisley\textsuperscript{\rm 2}}
\affiliations{
    \textsuperscript{\rm 1}South China University of Technoledge\\
    \textsuperscript{\rm 2}Columbia University
}

\usepackage{bibentry}
\begin{document}

\maketitle

\begin{abstract}
Bayesian Last Layer (BLL) models focus solely on uncertainty in the output layer of neural networks, demonstrating comparable performance to more complex Bayesian models. However, the use of Gaussian priors for last layer weights in  Bayesian Last Layer (BLL) models limits their expressive capacity when faced with non-Gaussian, outlier-rich, or high-dimensional datasets. To address this shortfall, we introduce a novel approach that combines diffusion techniques and implicit priors for variational learning of Bayesian last layer weights. This method leverages implicit distributions for modeling weight priors in BLL, coupled with diffusion samplers for approximating true posterior predictions, thereby establishing a comprehensive Bayesian prior and posterior estimation strategy. By delivering an explicit and computationally efficient variational lower bound, our method aims to augment the expressive abilities of BLL models, enhancing model accuracy, calibration, and out-of-distribution detection proficiency. Through detailed exploration and experimental validation, We showcase the method's potential for improving predictive accuracy and uncertainty quantification while ensuring computational efficiency.
\end{abstract}


\section{Introduction}
\label{intro}

Bayesian Last Layer (BLL) models \cite{watson2021latent,harrison2020meta,kristiadi2020being,harrison2023variational,fiedler2023improved} have emerged as a robust framework for uncertainty quantification in neural networks, concentrating on the uncertainty inherent in the final layer weights. However, these models often utilize Gaussian priors for the weight distributions, which may be insufficient for capturing the complexity of non-Gaussian, outlier-prone, or high-dimensional data. This constraint can limit the expressiveness of BLL models and adversely affect their performance in more challenging scenarios.

Prior research highlights the critical need to enhance model flexibility through more adaptable priors. For instance, \cite{fortuin2021bayesian} demonstrated that isotropic Gaussian priors may inadequately represent the true distribution of weights in Bayesian neural networks, potentially compromising performance. They observed significant spatial correlations in convolutional neural networks and ResNets, as well as heavy-tailed distributions in fully connected networks, suggesting that priors designed with these observations in mind can improve performance on image classification tasks. \cite{fortuin2022priors} underscore the importance of prior selection in Bayesian deep learning, exploring various priors for deep Gaussian processes \cite{damianou2013deep}, variational autoencoders \cite{doersch2016tutorial}, and Bayesian neural networks \cite{kononenko1989bayesian}, while also discussing methods for learning priors from data. Their work encourages practitioners to carefully consider prior specification and provides inspiration for this aspect. %\cite{tran2022all,tran2020functional} propose a framework for inference with functional priors, using Gaussian processes to rigorously define functional priors and facilitating the alignment of neural network parameter priors with the objectives of the functional prior.

Driven by the need for enhanced model flexibility and performance, we propose an innovative approach that leverages implicit priors for variational learning of Bayesian last layer weights. Implicit priors are parameterized through a neural network, replacing traditional Gaussian weight distributions to achieve greater flexibility. This method connects to the migration of variational implicit processes \cite{ma2019variational} into the BLL model, offering novel insights and opportunities. By employing implicit distributions for weight priors, our approach aims to establish a robust strategy for Bayesian prior estimation. However, as model complexity increases, inference becomes more challenging, posing new obstacles. 

To address this, we shift to directly utilizing diffusion models \cite{ho2020denoising,rombach2022high,vargas2023denoising} for posterior sampling. This approach enables us to effectively capture complex dependencies and correlations among latent variables. By utilizing diffusion stochastic differential equations (SDEs) and incorporating elements similar to score matching \cite{song2020score}, we formulate a novel variational lower bound for the marginal likelihood function through KL divergence minimization.

Additional, we delve into the details of our proposed method and demonstrate its potential through extensive experimental validation. By introducing a computationally efficient variational lower bound and showcasing its efficacy in scenarios with non-Gaussian distributions, outliers, and high-dimensional data, we highlight the significance of our approach in advancing uncertainty quantification in neural networks. Overall, our contributions are as follows:
\begin{itemize}
    \item We proposed an innovative approach that utilizes implicit priors for variational learning of Bayesian last layer weights. This method replaces traditional Gaussian weight parameters with prior distributions parameterized through a neural network to achieve more flexible priors. 
    
    \item We directly employed diffusion models for posterior sampling and then constructed a new objective that accurately captures the complex dependencies and correlations among latent variables. This approach explicitly derives a variational lower bound for the marginal likelihood function through KL divergence minimization.
    
    \item We conduct extensive experiments to demonstrate the effectiveness of the proposed method in handling regression and image classification datasets. These tests highlight the method's impact on improving uncertainty quantification in neural networks, demonstrating its effectiveness and robustness.
\end{itemize}

\section{Background}
\subsection{BLL models}
In the context of Bayesian Last Layer (BLL) networks, these models can be understood as Bayesian linear regression frameworks applied within a feature space that is adaptively learned through neural network architectures. Another perspective is to view them as neural networks where the parameters of the final layer undergo rigorous Bayesian inference. While BLL networks are capable of handling multivariate regression tasks, this discussion will be confined to the case of univariate targets for the sake of simplicity.

Let the dataset $\mathcal{D}=\left\{\left(\mathbf{x}_i, y_i\right)\right\}_{i=1}^N$ consist of observed data, where $\mathbf{x}_i \in \mathbb{R}^D$ and $y_i \in \mathbb{R}$. We denote $\mathbf{X} \in \mathbb{R}^{N \times D}$ and $\mathbf{y} \in \mathbb{R}^N$. We define a parameterized function $\boldsymbol{\phi}(\cdot ; \boldsymbol{\theta}): \mathbb{R}^D \rightarrow \mathbb{R}^M$ for projecting features, with $\boldsymbol{\theta}$ representing its parameters, and $\phi_i=\boldsymbol{\phi}\left(\mathbf{x}_i ; \boldsymbol{\theta}\right)$ and $\boldsymbol{\Phi}=\left[\boldsymbol{\phi}_1^{\top} \ldots \boldsymbol{\phi}_N^{\top}\right] \in \mathbb{R}^{N \times M}$, where the latter denotes a matrix of stacked row vectors.

A latent function $f$ is modeled utilizing Bayesian linear regression \cite{bishop2006pattern, hill1974bayesian,murphy2023probabilistic}, incorporating weights $\boldsymbol{\beta} \in \mathbb{R}^M $ and zero-mean Gaussian noise $\epsilon$ with variance $\sigma^2$, where
\begin{equation}
y_i=f\left(\mathbf{x}_i ; \boldsymbol{\theta}\right)=\boldsymbol{\phi}_i^{\top} \boldsymbol{\beta}+\epsilon_i .    
\end{equation}
By imposing a conjugate Gaussian prior $\mathcal{N}\left(\boldsymbol{\mu}_0, \boldsymbol{\Lambda}_0^{-1}\right)$ over $\boldsymbol{\beta}$, a Gaussian posterior $\mathcal{N}\left(\boldsymbol{\mu}_n, \boldsymbol{\Lambda}_n^{-1}\right)$ is obtained \cite{watson2021latent, harrison2023variational}, 
\begin{equation}
\begin{aligned}
\boldsymbol{\mu}_n &= (\boldsymbol{\Lambda}_0 + \sigma^{-2}\boldsymbol{\Phi}^T\boldsymbol{\Phi})^{-1}(\boldsymbol{\Lambda}_0\boldsymbol{\mu}_0 + \sigma^{-2}\boldsymbol{\Phi}^T\mathbf{y})\\
\boldsymbol{\Lambda}_n &= \boldsymbol{\Lambda}_0 + \sigma^{-2}\boldsymbol{\Phi}^T\boldsymbol{\Phi},
\end{aligned}
\end{equation}
yielding an explicit Gaussian predictive distribution for a given query $\mathbf{x}$,
\begin{equation}
    y \mid \mathbf{x}, \mathcal{D}, \boldsymbol{\theta} \sim \mathcal{N}\left(\cdot \mid \boldsymbol{\phi}_{\mathbf{x}}^{\top} \boldsymbol{\mu}_n, \sigma^2+\boldsymbol{\phi}_{\mathbf{x}}^{\top} \boldsymbol{\Lambda}_n^{-1} \boldsymbol{\phi}_{\mathbf{x}}\right),
\end{equation}
where $\boldsymbol{\mu}_n$ and $\boldsymbol{\Lambda}_n$ denote the mean vector and precision matrix of the posterior weight distribution, respectively.

The parameters encompassing the observation noise $\sigma^2$, the prior weight parameters $\boldsymbol{\mu}_0$ and $\boldsymbol{\Lambda}_0$, and $\boldsymbol{\theta}$, can be  optimized jointly through the maximization of the log-marginal likelihood. In the scenario where $\boldsymbol{\mu}_0=\mathbf{0}$, this model aligns equivalently with a Gaussian process, with a kernel $k\left(\mathbf{x}, \mathbf{x}^{\prime} ; \boldsymbol{\theta}\right)=\boldsymbol{\phi}(\mathbf{x} ; \boldsymbol{\theta})^{\top} \boldsymbol{\Lambda}_0^{-1} \boldsymbol{\phi}\left(\mathbf{x}^{\prime} ; \boldsymbol{\theta}\right)$ \cite{williams2006gaussian}. For a more rigorous Bayesian treatment, an inverse gamma prior can be placed on $\sigma^2$, eliciting a Student-$t$ weight posterior and predictive density. 

\subsection{Variational Bayesian Last Layer}

To leverage exact marginalization while avoiding the computational burden of full marginal likelihood computation, Variational Bayesian Last Layer (VBLL) \cite{harrison2023variational} employs stochastic variational inference \citep{hoffman2013stochastic}. The objective is to jointly compute an approximate posterior for the last layer parameters and optimize network weights by maximizing lower bounds on the marginal likelihood. Specifically, VBLL aims to find an approximate posterior \( q(\boldsymbol{\beta} | \eta) \) parameterized by \(\eta\).

Given a mini-batch \(\mathcal{D}_I\) with \(\left | \mathbf{X}_I\ \right | = B\), where \(I \subset \{1,2,\ldots,N\}\) is the set of indices for any mini-batch, and the log marginal likelihood \(\log p(y | \mathbf{x}, \theta)\) with marginalized parameters \(q(\boldsymbol{\beta} | \eta) = \mathcal{N}(\boldsymbol{w}, S)\), they derive bounds of the form:
\begin{equation}
\begin{aligned}
&\frac{N}{B} \log p(\mathbf{y}_I | \mathbf{X}_I, \theta) \geq \frac{N}{B} \sum_{i=1}^{B}[ \log \mathcal{N}(y_i | \boldsymbol{w}^\top \boldsymbol{\phi}_i, \sigma^2I) \\&- \frac{1}{2} \sigma^{-2} \boldsymbol{\phi}_t^\top S \boldsymbol{\phi}_t] - \frac{N}{B} \text{KL}(q(\boldsymbol{\beta} | \eta) \| p(\boldsymbol{\beta})),    
\end{aligned}
\end{equation}

This formulation results in a mini-batch algorithm for variational Bayesian inference in neural networks.

\subsection{Flexible Priors in Bayesian Neural Networks}
In prior research, there has been a strong emphasis on the importance of enhancing model flexibility through more flexible priors, particularly in the context of Bayesian neural networks. Previous studies \cite{fortuin2021bayesian,fortuin2022priors} have highlighted that applying isotropic Gaussian priors to weights may not fully capture true beliefs about weight distributions, thus hindering optimal performance, as 
described in Section \ref{intro} . 

These studies provide valuable insights suggesting that introducing more flexible weight priors in Bayesian linear regression models could enhance model flexibility and performance. This flexibility may involve designing prior distributions based on specific task or data characteristics to better capture the true data distribution features, thereby improving model generalization and performance. 

\section{Method}
\subsection{Implict Prior}
In continuation of techniques for generating implicit processes from prior research on stochastic processes \cite{ma2019variational,santana2021function,ma2021functional} , we introduce a regression model with an implicit prior over the weights $\boldsymbol{\beta}$:
\begin{align}
\label{eq4}
    \begin{aligned}
    &y_i=f\left(\mathbf{x}_i ; \boldsymbol{\theta}\right)=\boldsymbol{\phi}_i^{\top} \boldsymbol{\beta}+\epsilon_i,\\
    &\boldsymbol{\beta}=G_{\boldsymbol{\psi}}\left(\boldsymbol{\omega}\right), \boldsymbol{\omega} \sim p\left(\boldsymbol{\omega}\right),
    \end{aligned}
\end{align}
where $\boldsymbol{\omega} \in \mathbb{R}^K $ serves as an auxiliary variable to generate weights $\boldsymbol{\beta}$, with $p(\omega)$ representing its prior distribution, which can be a simple distribution such as a Gaussian distribution. And $G_{\boldsymbol{\psi}}(\cdot): \mathbb{R}^K \rightarrow \mathbb{R}^M$ being a neural network parameterized weight parameter generator. We use this hierarchical model to reframe the BLL model. Given that $G_{\boldsymbol{\psi}}$ is typically a non-linear function, we anticipate that the weight prior $p(\boldsymbol{\beta})$ generated by the auxiliary variable $\boldsymbol{\beta}$ through $G_{\boldsymbol{\psi}}$ will exhibit greater expressiveness than the original model, capable of yielding non-Gaussian distributions.

However, dealing with the intractable implicit model $p(\mathbf{y} \mid \mathbf{x}, \boldsymbol{\theta},\boldsymbol{\psi}, \boldsymbol{\omega})$ defined by Eq. (\ref{eq4}) poses significant challenges as model complexity increases, rendering traditional mean-field inference methods like those suggested in VBLL\cite{hoffman2013stochastic,harrison2023variational} inadequate for our modified model. As the complexity of the model increases, the posterior distribution becomes more intricate. Consequently, it is natural to design more sophisticated posteriors to model Bayesian networks and improve estimation accuracy. To address this issue, we explore an  approach centered on posterior sampling using diffusion models \cite{song2020score}. 
\subsection{Diffusion Posterior Sampling}

\subsubsection{Parameterizing Auxiliary Variable Posteriors Using Time-Reversal Representation of Diffusion SDE}

Our goal is to sample from the posterior distribution of auxiliary variable  \( q(\boldsymbol{\omega}) \), defined by Bayes' rule as \( q(\boldsymbol{\omega}) = \frac{p(\mathbf{y}|\boldsymbol{\omega})p(\boldsymbol{\omega})}{p(\mathbf{y})} \), where \( p(\boldsymbol{\omega}) \) represents the prior and \( p(\mathbf{y}) \) represents the model marginal likelihood. Following similar setups in prior works \cite{tzen2019theoretical, zhang2021path, vargas2023denoising}, we begin by sampling from a Gaussian distribution \( \mathcal{N}(0, \sigma_0^2I) \), where \( \sigma_0 \in \mathbb{R}^+ \) is the covariance parameter. We then follow a time-reversal process of the forward diffusion stochastic differential equation (SDE):
\begin{equation}
\label{eq2}
\mathrm{d}\overrightarrow{\boldsymbol{\omega}}_t = -\mathbf{\lambda}(t) \overrightarrow{\boldsymbol{\omega}}_t \mathrm{d}t + g(t)\mathrm{d}B_t, \quad \overrightarrow{\boldsymbol{\omega}}_0 \sim q, \quad t \in [0, T],
\end{equation}
where \( -\mathbf{\lambda}(t) \in \mathbb{R} \) is the drift coefficient, \( g(t) \in \mathbb{R} \) is the diffusion coefficient, and \( (B_t)_{t \in [0, T]} \) is a \( K \)-dimensional Brownian motion. This diffusion induces the path measure \( \mathcal{P} \) on the time interval \( [0, T] \), and the marginal density of \( \overrightarrow{\boldsymbol{\omega}}_t \) is denoted \( p_t \). Note that by definition, we always have \( p_0 = q \) when using an SDE to perturb this distribution. According to \cite{anderson1982reverse, haussmann1986time}, the time-reversal representation of Eq. (\ref{eq2}) is given by \( \overleftarrow{\boldsymbol{\omega}}_t = \overrightarrow{\boldsymbol{\omega}}_{T-t} \) (where equality is in distribution). This satisfies:
\begin{equation}
\label{eq3}
\begin{aligned}
&\mathrm{d} \overleftarrow{\boldsymbol{\omega}}_t = \left( -\lambda(T-t) \overleftarrow{ \boldsymbol{\omega}}_t + g(T-t)^2 \nabla \ln p_{T-t}\left( \overleftarrow{\boldsymbol{\omega}}_t \right) \right) \mathrm{d} t \\&+ g(T-t) \mathrm{d} W_t, \quad \overleftarrow{\boldsymbol{\omega}}_0 \sim p_T,
\end{aligned}
\end{equation}
where \( (W_t)_{t \in [0, T]} \) is another \( K \)-dimensional Brownian motion. In DDPM \cite{ho2020denoising, song2020score}, this time-reversal starts from \( \overleftarrow{\boldsymbol{\omega}}_0 \sim p_T \approx \mathcal{N}(0, \sigma_0^2 I) \) and ensures that \( \overleftarrow{\boldsymbol{\omega}}_T \sim q \). Then we can parameterize the transition probability \( \mathcal{T}\left( \boldsymbol{\omega}_{t_s+1} \mid \boldsymbol{\omega}_{t_s} \right) \) in the Euler discretized form \cite{sarkka2019applied} of Eq. (\ref{eq3}) for steps \( t_s \in \{0, \ldots, T-1\} \).

\subsubsection{Score Matching and Reference Process Trick}
If we could approximately simulate the diffusion process described in (\ref{eq3}), we would obtain approximate samples from the target distribution \(q\). However, implementing this idea requires approximating the intractable scores \(\left(\nabla \ln p_t(\cdot)\right)_{t \in[0, T]}\). To achieve this, DDPM \cite{ho2020denoising,song2020score} relies on score matching techniques. Specifically, to approximate \(\mathcal{P}\), we consider a path measure \(\mathcal{P}^{\boldsymbol{\gamma}}\) whose time-reversal is defined by
\begin{equation}
\label{eq5}
\begin{aligned}
&\mathrm{d} \overleftarrow{\boldsymbol{\omega}}_t^{\boldsymbol{\gamma}}=\left(-\mathbf{\lambda}(T-t)\overleftarrow{\boldsymbol{\omega}}_t^{\boldsymbol{\gamma}}+g(T-t)^2 s_{\boldsymbol{\gamma}}\left(T-t, \overleftarrow{\boldsymbol{\omega}}_t^{\boldsymbol{\gamma}}\right)\right) \mathrm{d} t\\&+g(T-t) \mathrm{d} W_t, \quad \overleftarrow{\boldsymbol{\omega}}_0^{\boldsymbol{\gamma}} \sim \mathcal{N}(0, \sigma_0^2 I),     
\end{aligned}
\end{equation}
so that the backward process \(\overleftarrow{\boldsymbol{\omega}}_t^{\boldsymbol{\gamma}} \sim \mathcal{Q}_{t}^{\boldsymbol{\gamma}}\). To obtain \(s_{\boldsymbol{\gamma}}(t, \cdot) \approx \nabla \ln p_t(\cdot)\), we parameterize \(s_{\boldsymbol{\gamma}}(t, \cdot)\) using a neural network, with the parameters obtained by minimizing \(\mathrm{KL}(\mathcal{P}||\mathcal{P}^{\boldsymbol{\gamma}})\). Unlike traditional score matching techniques, given that we can only obtain samples from \(\mathcal{Q}_{t}^{\boldsymbol{\gamma}}\), we alternatively minimize \(\mathrm{KL}(\mathcal{P}^{\boldsymbol{\gamma}}||\mathcal{P})\), by Girsanov's theorem \cite{oksendal2013stochastic}
\begin{equation}
\label{eq7}
\begin{aligned}
&\mathrm{KL}(\mathcal{P}^{\boldsymbol{\gamma}}||\mathcal{P})=\mathrm{KL}(\mathcal{Q}^{\boldsymbol{\gamma}}||\mathcal{Q})\\
=&\mathrm{KL}(\mathcal{N}(0, \sigma_0^2 I)||p_T)+\mathrm{KL}(\mathcal{Q}^{\boldsymbol{\gamma}}(\cdot|\overleftarrow{\boldsymbol{\omega}}_0^{\boldsymbol{\gamma}})||\mathcal{Q}(\cdot|\overleftarrow{\boldsymbol{\omega}}_0))\\
\approx &\frac{1}{2} \int_0^T \mathbb{E}_{\mathcal{Q}_t^{\boldsymbol{\gamma}}}[g(T-t)^2\|\nabla \ln p_{T-t}(\overleftarrow{\boldsymbol{\omega}}_t^{\boldsymbol{\gamma}})-\boldsymbol{s}_{\boldsymbol{\gamma}}(T-t, \overleftarrow{\boldsymbol{\omega}}_t^{\boldsymbol{\gamma}})\|_2^2] \mathrm{d} t
\end{aligned}
\end{equation}
However, although we can obtain samples from \(\mathcal{Q}_{t}^{\boldsymbol{\gamma}}\) by simulating the SDE (\ref{eq5}), dealing with the nonlinear drift function of SDE (\ref{eq5}) makes it difficult to obtain \(\nabla \ln p_{T-t}(\overleftarrow{\boldsymbol{\omega}}_t^{\boldsymbol{\gamma}})\) in Eq. (\ref{eq7}).

We use an alternative approach by constructing a reference process \cite{zhang2021path,vargas2023denoising}, denoted as \(\mathcal{P}^\mathrm{ref}\), to assist in measuring \(\mathrm{KL}(\mathcal{P}^{\boldsymbol{\gamma}}||\mathcal{P})\). Firstly, observe the following fact:
\begin{equation}
\label{eq8}
\begin{aligned}
\mathrm{KL}(\mathcal{P}^{\boldsymbol{\gamma}}||\mathcal{P}) &= \mathbb{E}_{\mathcal{P}^{\boldsymbol{\gamma}}} \log \frac{\mathrm{d}\mathcal{P}^{\boldsymbol{\gamma}}}{\mathrm{d}\mathcal{P}} \\
&= \mathbb{E}_{\mathcal{P}^{\boldsymbol{\gamma}}} \log \frac{\mathrm{d}\mathcal{P}^{\boldsymbol{\gamma}}}{\mathrm{d}\mathcal{P}^{\mathrm{ref}}} + \mathbb{E}_{\mathcal{P}^{\boldsymbol{\gamma}}} \log \frac{\mathrm{d}\mathcal{P}^{\mathrm{ref}}}{\mathrm{d}\mathcal{P}},
\end{aligned}
\end{equation}
where the stochastic process \(\mathrm{KL}\) is represented as the Radon-Nikodym derivative. Given the specific form in Eq. (\ref{eq8}), we define the reference process \(\mathcal{P}^\mathrm{ref}\) to follow the diffusion formula as in Eq. (\ref{eq2}), but initialized at \(p_0^{\mathrm{ref}}(\overrightarrow{\boldsymbol{\omega}}_0^\mathrm{ref}) = \mathcal{N}(0, \sigma_0^2 I)\) instead of \(q\), which aligns with the distribution of \(\overleftarrow{\boldsymbol{\omega}}_0\) in Eq. (\ref{eq5}),
\begin{equation}
\label{eq9}
   \mathrm{d}\overrightarrow{\boldsymbol{\omega}}_t^\mathrm{ref} = \mathbf{\lambda}(t)\overrightarrow{\boldsymbol{\omega}}_t^\mathrm{ref} \mathrm{d}t + g(t) \mathrm{d}B_t, \quad \overrightarrow{\boldsymbol{\omega}}_0^\mathrm{ref} \sim \mathcal{N}(0, \sigma_0^2 I).
\end{equation}
The transition kernel \(p_t(\overrightarrow{\boldsymbol{\omega}}_t^\mathrm{ref}|\overrightarrow{\boldsymbol{\omega}}_0^\mathrm{ref})\) is always a Gaussian distribution \(\mathcal{N}(l_t, \Sigma_t)\), where the mean \(l_t\) and variance \(\Sigma_t\) are often available in closed form \cite{sarkka2019applied}:
\begin{equation}
\label{eq10}
\begin{aligned}
\frac{\mathrm{d} l_t}{\mathrm{d} t} &= -\lambda(t) l_t, \quad l_0 = 0, \\
\frac{\mathrm{d} \Sigma_t}{\mathrm{d} t} &= -2 \lambda(t) \Sigma_t + g(t)^2 I, \quad \Sigma_0 = \sigma_0^2 I.
\end{aligned}
\end{equation}
By solving these ordinary differential equations \cite{hale2013introduction}, we obtain the general solutions as follows:
\begin{equation}
\label{eq11}
\begin{aligned}
l_t &= l_0 e^{-\int_0^t \lambda(s) \mathrm{d}s}, \\
\Sigma_t &= \left(\int_0^t g(r)^2 e^{\int_0^r \lambda(s) \mathrm{d}s} \mathrm{d}r I + \Sigma_0\right) e^{-\int_0^t \lambda(s) \mathrm{d}s}.
\end{aligned}
\end{equation}
According to Eq. (\ref{eq11}), we can derive that for any \(t\), the distribution \(p_t^{\mathrm{ref}}\) of \(\overrightarrow{\boldsymbol{\omega}}_t^\mathrm{ref}\) is a zero-mean Gaussian distribution:
\begin{equation}
\label{eq12}
\begin{aligned}
p_t^{\mathrm{ref}}(\overrightarrow{\boldsymbol{\omega}}_t^\mathrm{ref}) &= \int p_t(\overrightarrow{\boldsymbol{\omega}}_t^\mathrm{ref}|\overrightarrow{\boldsymbol{\omega}}_0^\mathrm{ref}) p_t(\overrightarrow{\boldsymbol{\omega}}_0^\mathrm{ref}) \, \mathrm{d}\overrightarrow{\boldsymbol{\omega}}_0^\mathrm{ref} \\
&= \mathcal{N}(0, \kappa_t I),
\end{aligned}
\end{equation}
where the variance \(\kappa_t\) is given by
\[
\kappa_t = \left(\int_0^t g(r)^2 e^{\int_0^r \lambda(s) \mathrm{d}s} \mathrm{d}r + \sigma_0^2\right) e^{-\int_0^t \lambda(s) \mathrm{d}s}.
\]
Meanwhile, the SDE equation for the reverse process \(\mathcal{Q}^\mathrm{ref}\) of \(\mathcal{P}^\mathrm{ref}\) is
\begin{equation}
\label{eq13}
\begin{aligned}
\mathrm{d} \overleftarrow{\boldsymbol{\omega}}_t^\mathrm{ref} &= \left(-\mathbf{\lambda}(T-t) \overleftarrow{\boldsymbol{\omega}}_t^\mathrm{ref} + g(T-t)^2 \nabla \ln p_{T-t}^\mathrm{ref} \left(\overleftarrow{\boldsymbol{\omega}}_t^\mathrm{ref}\right)\right) \mathrm{d}t \\&+ g(T-t) \mathrm{d}W_t, \quad \overleftarrow{\boldsymbol{\omega}}_0^\mathrm{ref} \sim p_T^\mathrm{ref}.
\end{aligned}
\end{equation}
According to Eq. (\ref{eq12}), we can derive an analytical expression for the derivative of the log-likelihood function with respect to \(\overleftarrow{\boldsymbol{\omega}}_t^\mathrm{ref}\):
\begin{equation}
\label{eq14}
    \nabla \ln p_{T-t}^\mathrm{ref}\left(\overleftarrow{\boldsymbol{\omega}}_t^\mathrm{ref}\right) = -\frac{\overleftarrow{\boldsymbol{\omega}}_t^\mathrm{ref}}{\kappa_{T-t}}.
\end{equation}
To compute \(\mathrm{KL}(\mathcal{P}^{\boldsymbol{\gamma}}||\mathcal{P})\), we calculate the first term of Eq. (\ref{eq8}). Using the chain rule for KL divergence and Girsanov's theorem \cite{oksendal2013stochastic}, and incorporating Eqs. (\ref{eq5}, \ref{eq13}, \ref{eq14}), we obtain:
\begin{equation}
\label{eq15}
\begin{aligned}
 &\mathbb{E}_{\mathcal{P}^{\boldsymbol{\gamma}}} \log \frac{\mathrm{d} \mathcal{P}^{\boldsymbol{\gamma}}}{\mathrm{d} \mathcal{P}^{\mathrm{ref}}}= \mathrm{KL}\left(\mathcal{P}^{\boldsymbol{\gamma}} \| \mathcal{P}^{\mathrm{ref}}\right) = \mathrm{KL}\left(\mathcal{Q}^{\boldsymbol{\gamma}} \| \mathcal{Q}^{\mathrm{ref}}\right) 
\\=& \mathrm{KL}\left(\mathcal{N}(0, \sigma_0^2 I) \| p_T^{\mathrm{ref}}\right) + \mathrm{KL}\left(\mathcal{Q}^{\boldsymbol{\gamma}}(\cdot|\overleftarrow{\boldsymbol{\omega}}_0^{\boldsymbol{\gamma}})\| \mathcal{Q}(\cdot|\overleftarrow{\boldsymbol{\omega}}_0^\mathrm{ref})\right) 
\\=&\mathrm{KL}(\mathcal{N}(0, \sigma_0^2 I) \| p_T^{\mathrm{ref}}) \\+& \frac{1}{2} \int_0^T \mathbb{E}_{\mathcal{Q}_t^{\boldsymbol{\gamma}}} \left[g(T-t)^2 \left\|\frac{\overleftarrow{\boldsymbol{\omega}}_t^{\boldsymbol{\gamma}}}{\kappa_{T-t}} + \boldsymbol{s}_{\boldsymbol{\gamma}}(T-t, \overleftarrow{\boldsymbol{\omega}}_t^{\boldsymbol{\gamma}})\right\|_2^2\right] \mathrm{d}t.
\end{aligned}
\end{equation}
At this point, we can simulate the SDE (\ref{eq5}) to compute the first term in Eq. (\ref{eq8}).  the integral term can be computed using either ODE solvers \cite{chen2018neural} or by employing Riemann summation methods. For the second term, $\mathbb{E} _{\mathcal{P} ^{{\boldsymbol{\gamma}}}}\log \frac{\mathrm{d}\mathcal{P} ^{\mathrm{ref}}}{\mathrm{d}\mathcal{P}}$, we can see from Eq. (\ref{eq2}) and Eq. (\ref{eq9}) that $\mathcal{P}$ and $\mathcal{P}^{\mathrm{ref}}$ have the same dynamic system $\tau$, except for different initial values. Therefore, we have

\begin{equation}
\label{eq16}
\begin{aligned}
\mathbb{E} _{\mathcal{P} ^{{\boldsymbol{\gamma}}}}\log \frac{\mathrm{d}\mathcal{P} ^{\mathrm{ref}}}{\mathrm{d}\mathcal{P}}&=\mathbb{E} _{\mathcal{P} ^{{\boldsymbol{\gamma}}}}\log \frac{\mathcal{P} ^{\mathrm{ref}}\left( \tau |\cdot \right) p_{0}^{\mathrm{ref}}\left( \cdot \right)}{\mathcal{P} \left( \tau |\cdot \right) p_0\left( \cdot \right)}
\\&=\mathbb{E} _{\mathcal{Q}_T ^{{\boldsymbol{\gamma}}}}\log \frac{p_{0}^{\mathrm{ref}}\left( \cdot \right)}{p_0\left( \cdot \right)}=\mathbb{E} _{\mathcal{Q}_T ^{{\boldsymbol{\gamma}}}}\log \frac{\mathcal{N}(0, \sigma_0^2I)}{q}
\\
&=\mathbb{E} _{\mathcal{Q}_T ^{{\boldsymbol{\gamma}}}}\log \frac{\mathcal{N}(0, \sigma_0^2I)}{p(\mathbf{y}|\cdot)p(\cdot)}+\log p(\mathbf{y})
\end{aligned}
\end{equation}
\begin{table*}[ht]
\caption{Results for UCI regression tasks. 
}\label{tab:reg1results}

\centering
\resizebox{\textwidth}{!}{
\begin{tabular}{c|cc|cc|cc} 
& \multicolumn{2}{c}{\textsc{Boston}} & \multicolumn{2}{c}{\textsc{Concrete}} & \multicolumn{2}{c}{\textsc{Energy}}\\

& NLL ($\downarrow$) & RMSE ($\downarrow$) & NLL ($\downarrow$) & RMSE ($\downarrow$) & NLL ($\downarrow$) & RMSE ($\downarrow$) \\ 

\hline

VBLL & ${2.55 \pm 0.06}$ & $2.92 \pm 0.12$ & ${3.22 \pm 0.07}$ & ${5.09 \pm 0.13}$ & ${1.37 \pm 0.08}$ & ${0.87 \pm 0.04}$\\ 
GBLL & $2.90\pm0.05$ & $4.19\pm0.17$ & $3.09 \pm 0.03$ & $5.01 \pm 0.18$ & $0.69 \pm 0.03$ & $0.46 \pm 0.02$ \\ 
LDGBLL & $2.60\pm0.04$ & $3.38\pm0.18$ & $2.97 \pm 0.03$ & $4.80 \pm 0.18$ & $4.80 \pm 0.18$ & $0.50 \pm 0.02$ \\ 
MAP & $2.60\pm0.07$ & $3.02\pm0.17$ & $3.04 \pm 0.04$ & $4.75 \pm 0.12$ & $1.44 \pm 0.09$ & $0.53 \pm 0.01$\\ 
RBF GP & $2.41\pm0.06$ & $2.83 \pm 0.16$ & $3.08 \pm 0.02$ & $5.62 \pm 0.13$ & $0.66 \pm 0.04$ & $0.47 \pm 0.01$\\ 
\hline
Dropout & $2.36\pm0.04$ & $2.78\pm0.16$ & $2.90 \pm 0.02$ & $4.45 \pm 0.11$ & $1.33 \pm 0.00$ & $0.53 \pm 0.01$ \\ 
Ensemble & $2.48\pm0.09$ & $2.79\pm0.17$ & $3.04 \pm 0.08$ & $4.55 \pm 0.12$ & $\bm{0.58 \pm 0.07}$ & $\bm{0.41 \pm 0.02}$\\ 
SWAG & $2.64\pm0.16$ & $3.08\pm0.35$ & $3.19 \pm 0.05$ & $5.50 \pm 0.16$ & $1.23 \pm 0.08$ & $0.93 \pm 0.09$ \\ 
BBB & $2.39\pm0.04$ & $2.74\pm0.16$ & $2.97 \pm 0.03$ & $4.80 \pm 0.13$ & $0.63 \pm 0.05$ & ${0.43 \pm 0.01}$ \\ 

\hline
 DVI-IBLL (ours) & $\bm{2.12 \pm 0.05}$ & $\bm{2.49 \pm 0.10}$ & $\bm{2.66 \pm 0.04}$ & $\bm{4.08 \pm 0.11}$ & $1.19 \pm 0.11$ & $0.71 \pm 0.03$\\ 
\hline

\end{tabular}
}
\end{table*}

\begin{table*}[ht]
\caption{Further results for UCI regression tasks.}\label{tab:reg2results}
\centering
\resizebox{\textwidth}{!}{
\begin{tabular}{c|cc|cc|cc} 
& \multicolumn{2}{c}{\textsc{Power}} & \multicolumn{2}{c}{\textsc{Wine}} & \multicolumn{2}{c}{\textsc{Yacht}}\\

& NLL ($\downarrow$) & RMSE ($\downarrow$) & NLL ($\downarrow$) & RMSE ($\downarrow$) & NLL ($\downarrow$) & RMSE ($\downarrow$)\\ 

\hline

VBLL & $2.73 \pm 0.01$ & $3.68 \pm 0.03$ & $1.02 \pm 0.03$ & $0.65 \pm 0.01$ & $1.29 \pm 0.17$ & $0.86 \pm 0.17$\\ 
GBLL & ${2.77 \pm 0.01}$ & ${3.85 \pm 0.03}$ & $1.02 \pm 0.01$ & $0.64 \pm 0.01$ & $1.67 \pm 0.11$ & $1.09 \pm 0.09$\\ 
LDGBLL & $2.77 \pm 0.01$ & $3.85 \pm 0.04$ & $1.02 \pm 0.01$ & $0.64 \pm 0.01$ & $1.13 \pm 0.06$ & $0.75 \pm 0.10$\\ 
MAP & $2.77 \pm 0.01$ & $3.81 \pm 0.04$ & $0.96 \pm 0.01$ & $0.63 \pm 0.01$ & $5.14 \pm 1.62$ & $0.94 \pm 0.09$\\ 
RBF GP & $2.76 \pm 0.01$ & $3.72 \pm 0.04$ & $\bm{0.45 \pm 0.01}$ & $\bm{0.56 \pm 0.05}$ & $\bm{0.17 \pm 0.03}$ & $\bm{0.40 \pm 0.03}$\\ 
\hline
Dropout & $2.80 \pm 0.01$ & $3.90 \pm 0.04$ & $0.93 \pm 0.01$ & $0.61 \pm 0.01$ & $1.82 \pm 0.01$ & $1.21 \pm 0.13$\\ 
Ensemble & $2.70 \pm 0.01$ & $3.59 \pm 0.04$ & $0.95 \pm 0.01$ & $0.63 \pm 0.01$ & $0.35 \pm 0.07$ & $0.83 \pm 0.08$\\ 
SWAG & $2.77 \pm 0.02$ & $3.85 \pm 0.05$ & $0.96 \pm 0.03$ & $0.63 \pm 0.01$ & $1.11 \pm 0.05$ & $1.13 \pm 0.20$\\ 
BBB & $2.77 \pm 0.01$ & $3.86 \pm 0.04$ & $0.95 \pm 0.01$ & $0.63 \pm 0.01$ & $1.43 \pm 0.17$ & $1.10 \pm 0.11$\\ 

\hline
DVI-IBLL (ours) & \bm{$2.67 \pm 0.01$} & \bm{$3.58 \pm 0.03$} & $0.90 \pm 0.03$ & $0.60 \pm 0.01$ & $0.92 \pm 0.15$ & $0.76 \pm 0.09$\\ 
\hline
\end{tabular}
}

\end{table*}

\subsubsection{Evidence Lower Bound}

Let $l_1({\boldsymbol{\gamma}})=\mathbb{E} _{\mathcal{P} ^{{\boldsymbol{\gamma}}}}\log \frac{\mathrm{d}\mathcal{P} ^{{\boldsymbol{\gamma}}}}{\mathrm{d}\mathcal{P}^{\mathrm{ref }}}$. Combining Eq. (\ref{eq4}, \ref{eq8}, \ref{eq15}, \ref{eq16}), we obtain a new variational lower bound $l({\boldsymbol{\gamma}},\boldsymbol{\theta},\boldsymbol{\psi})$ for the marginal likelihood $\log p(\mathbf{y})$ in our method,
\begin{equation}
\label{eq17}
\begin{aligned}
&\log p(\mathbf{y})\\&= \mathrm{KL(}\mathcal{P} ^{{\boldsymbol{\gamma}}}||\mathcal{P} )-l_1({\boldsymbol{\gamma}})-\mathbb{E} _{\mathcal{Q}_T ^{{\boldsymbol{\gamma}}}}\log \frac{\mathcal{N}(0, \sigma_0^2I)}{p(\mathbf{y}|\mathbf{\cdot})p(\mathbf{\cdot})}\\
&= \mathrm{KL(}\mathcal{P} ^{{\boldsymbol{\gamma}}}||\mathcal{P} )-l_1({\boldsymbol{\gamma}})-\mathbb{E} _{\mathcal{Q}_T ^{{\boldsymbol{\gamma}}}}\log {\mathcal{N}(0, \sigma_0^2I)}+\mathbb{E} _{\mathcal{Q}_T ^{{\boldsymbol{\gamma}}}}\log p(\cdot)\\&+\mathbb{E} _{\mathcal{Q}_T^{{\boldsymbol{\gamma}}}}\log p(\mathbf{y} \mid \mathbf{x}, \boldsymbol{\theta},\boldsymbol{\psi}, \cdot)\\
&\ge -l_1({\boldsymbol{\gamma}})-\mathbb{E} _{\mathcal{Q}_T ^{{\boldsymbol{\gamma}}}}\log {\mathcal{N}(0, \sigma_0^2I)}+\mathbb{E} _{\mathcal{Q}_T ^{{\boldsymbol{\gamma}}}}\log p(\cdot)\\&+\mathbb{E} _{\mathcal{Q}_T^{{\boldsymbol{\gamma}}}}\log p(\mathbf{y} \mid \mathbf{x}, \boldsymbol{\theta},\boldsymbol{\psi}, \cdot)\\
&=l({\boldsymbol{\gamma}},\boldsymbol{\theta},\boldsymbol{\psi})
\end{aligned}
\end{equation}
In our derivation, \( p(\cdot) \) represents the prior function of \(\boldsymbol{\omega}\). We introduce a new variational lower bound for \(\log p(\mathbf{y})\). Unlike the mean-field variational inference model that approximates \( q \) with a Gaussian distribution, our model uses a diffusion process to approximate the posterior distribution. The flexibility of the denoising neural network \(\boldsymbol{\gamma}\) provides our model with a significant advantage in accurately approximating the posterior distribution.
\subsection{ Stochastic Gradient Descent}
\begin{algorithm*}[ht]
\caption{Diffuison
Variational Inference algorithm  for implict prior BLLs (\textbf{DVI-IBLL}) }
\label{algorithm}
\begin{algorithmic}
   \STATE {\bfseries Input:} training data $\mathbf{x}, \mathbf{y}$
   mini-batch size $B$
   \STATE {\bfseries Initialize}  diffusion coefficient $ h,g$, all BLL hyperparameters $\boldsymbol{\theta},\boldsymbol{\psi}$, denoising diffusion network parameters $\boldsymbol{\gamma}$
   \STATE Set $l_0=0$
   \REPEAT

   \FOR{$t_s=0$ {\bfseries to} $T-1$}
   
   \STATE Draw $\boldsymbol{\epsilon} _{t_s}$ from  standard Gaussian distribution.

   \STATE Set $\boldsymbol{\omega}_{t_s+1}=\boldsymbol{\omega}_{t_s}-\mathbf{\lambda}(\boldsymbol{T-t_s)\omega}_{t_s}+g(T-t_s)^2 s_{\boldsymbol{\gamma}} \left(T-t_s,\boldsymbol{\omega}_{t_s}\right)+g(T-t) \boldsymbol{\epsilon}_{t_s}$
   \STATE Compute $\kappa_{T-(t_s+1)}$ by Eq. (\ref{eq12})

   \STATE Set $l_{t_s+1}=l_{t_s}+g(T-(t_s+1))^2\|\frac{\boldsymbol{\omega}_{t_s+1}}{\kappa _{T-(t_s+1)}}+\boldsymbol{s}_{\boldsymbol{\gamma}}(T-(t_s+1), \boldsymbol{\omega}_{t_s+1})\|_2^2$
  
   \ENDFOR
   \STATE Sample  mini-batch indices $I \subset\{1, \ldots, N\} \text { with }|I|=B$ 
 
\STATE Set $l(\boldsymbol{\gamma},\boldsymbol{\theta},\boldsymbol{\psi})=\frac{1}{2}\left(\boldsymbol{\omega}_{T}^2\right)+B\log \sigma_0+\log p\left(\boldsymbol{\omega}_{T}\right)+\frac{N}{B}\log p(\mathbf{y}_I \mid \mathbf{x}_I, \boldsymbol{\theta},\boldsymbol{\psi},\boldsymbol{\omega}_{T})-{\mathrm{KL}\left( \mathcal{N}(0, \sigma_0^2I)\parallel \mathcal{N}(0, \kappa_T) \right)}-\frac{1}{2}l_T$ 
\STATE Do gradient descent on $l(\boldsymbol{\gamma},\boldsymbol{\theta},\boldsymbol{\psi})$
\UNTIL{$\boldsymbol{\gamma},\boldsymbol{\theta},\boldsymbol{\psi}$  converge}
\end{algorithmic}
\end{algorithm*}
For ease of sampling, we consider a reparameterization version of Eq. (\ref{eq17}) based on the approaximate transition probability $\mathcal{T}_{\boldsymbol{\gamma}}\left(\boldsymbol{\omega}_{t_s+1} \mid \boldsymbol{\omega}_{t_s}\right)$ given by
\begin{align}
\begin{aligned}
&\mathcal{T}_{\boldsymbol{\gamma} }(\boldsymbol{\omega}_{t_s+1})=\boldsymbol{\omega}_{t_s}-\mathbf{\lambda}(T-t_s)\boldsymbol{\omega}_{t_s}+g(T-t_s)^2 s_{\boldsymbol{\gamma}} \left(T-t_s,\boldsymbol{\omega}_{t_s}\right)\\&+g(T-t) \boldsymbol{\epsilon}_{t_s}. 
\end{aligned}
\end{align}
where $\boldsymbol{\epsilon}_{t_s} \sim \mathcal{N}(0,I)$. In order to accelerate training and sampling in our inference scheme, we propose a  scalable variational bounds that are tractable in the large data regime based on stochastic variational inference \cite{kingma2013auto,hoffman2015structured,salimbeni2017doubly,naesseth2020markovian} and stochastic gradient descent \cite{welling2011bayesian,chen2014stochastic,zou2019stochastic, alexos2022structured}. Our model is shown  in Algorithm \ref{algorithm}, referred to as \textbf{DVI-IBLL}.

\subsection{Prediction Distribution}

For making predictions in our model, the prediction under the variational posterior distribution is approximated for a test input/label $(\mathbf{x}^\star, \mathbf{y}^\star)$ as:

\begin{equation}
 p(\mathbf{y}^\star \mid \mathbf{x}^\star, \mathbf{X}, \mathbf{y}) \approx  \mathbb{E}_{\mathcal{Q}_T ^{\boldsymbol{\gamma}}}\left[ p(\mathbf{y}^\star \mid \mathbf{x}^\star, \boldsymbol{\theta},\boldsymbol{\psi}, \boldsymbol{\omega})\right]   
\end{equation}

Here, $\mathcal{Q}_T ^{\boldsymbol{\gamma}}$ denotes the output of the diffusion process at time $T$. The expression $p(\mathbf{y}^\star \mid \mathbf{x}^\star, \boldsymbol{\theta},\boldsymbol{\psi}, \boldsymbol{\omega})$ can be obtained by substituting the input/output with the test set input/label from Eq. (\ref{eq4}). For classification tasks or other likelihood functions, the substitution can be made accordingly during training.

\section{Related Work}

\subsection{Bayesian Last Layers (BLL) Models}
Bayesian Last Layers (BLL) models are a class of methods that enhance neural network performance by incorporating Bayesian principles into the final layers of the network. The primary advantage of BLL models lies in their ability to efficiently balance exploration and exploitation. Early work by \cite{box2011bayesian} integrated Bayesian layers with deep neural networks to improve robustness and generalization. Recent advances have further refined these approaches. For instance, \cite{weber2018optimizing} explored training neural networks online in a bandit setting to optimize the balance between exploration and exploitation. Additionally, \cite{watson2021latent} introduced a functional prior on the model’s derivatives with respect to the inputs, enhancing predictive uncertainty. \cite{harrison2023variational} applied variational inference to train Bayesian last layer neural networks, improving the estimation of posterior distributions. Moreover, \cite{fiedler2023improved} addressed computational challenges in the log marginal likelihood by reintroducing the weights of the last layer, avoiding the need for matrix inversion.

\subsection{Implicit Prior Models}
Implicit prior models \cite{hoffman2017beta, ma2019variational} in Bayesian inference refer to models where the prior distribution is not explicitly specified but instead learned through neural networks. These models are gaining traction due to their flexibility and ability to capture complex data distributions. Notably, \cite{ma2019variational} proposed highly flexible implicit priors over functions, exemplified by data simulators, Bayesian neural networks, and non-linear transformations of stochastic processes. \cite{takahashi2019variational} introduced the VAE with implicit optimal priors to address the challenges of hyperparameter tuning for the aggregated posterior model. Recent advancements \cite{kumar2020implicit} have focused on enhancing the efficiency and accuracy of these methods by integrating regularization techniques.

\subsection{Diffusion Models}
Diffusion models \cite{ho2020denoising, song2020score} have emerged as powerful tools for modeling complex distributions and generating high-quality samples. These models simulate a diffusion process in which a simple distribution is gradually transformed into a more complex target distribution. The incorporation of stochastic differential equations (SDEs) \cite{song2020score} has further enhanced their capacity to model continuous dynamical systems and capture intricate data patterns. Recent research \cite{vargas2023denoising, richter2023improved, piriyakulkij2024diffusion} has explored integrating diffusion models with Bayesian inference, resulting in the generation of unnormalized probability density function (PDF) samples. Our approach bears significant resemblance to \cite{vargas2023denoising}, but it is distinct in that it derives a different form of SDE aimed at posterior sampling. Unlike \cite{piriyakulkij2024diffusion}, our method does not require the introduction of additional auxiliary variables or wake-sleep algorithms, enabling end-to-end optimization.

\begin{table*}[ht]
\caption{Results for Wide ResNet-28-10 on CIFAR-10.}\label{tab:cifar10results}
\centering
% \scriptsize
\resizebox{\linewidth}{!}{
\begin{tabular}{c|ccccc} 
Method & Accuracy ($\uparrow$) & ECE ($\downarrow$) & NLL ($\downarrow$) & SVHN AUC ($\uparrow$) & CIFAR-100 AUC ($\uparrow$)              \\ 
\hline
DNN   & $95.8 \pm 0.19$  &  $0.028 \pm 0.028$ & $0.183 \pm 0.007$ & $0.946 \pm 0.005$ & $0.893 \pm 0.001$ \\ 
% \hline
SNGP & $95.7 \pm 0.14$  & ${0.017 \pm 0.003}$ & ${0.149 \pm 0.005}$ & $0.960 \pm 0.004$ & $0.902 \pm 0.003$ \\ 
% \hline
D-VBLL & $96.4 \pm 0.12$ & $0.022 \pm 0.001$ &  $0.160 \pm 0.001$  & $0.969 \pm 0.004$ & $0.900 \pm 0.004$\\
G-VBLL & $96.3 \pm 0.06$  & $0.021 \pm 0.001$ & $0.174 \pm 0.002$ & $0.925 \pm 0.015$ & $0.804 \pm 0.006$   \\         
\hline
DNN + LL Laplace & $96.3 \pm 0.03$ & $\bm{0.010 \pm 0.001}$ & $0.133 \pm 0.003$ & $0.965 \pm 0.010$ & $0.898 \pm 0.001$\\
DNN + D-VBLL & $96.4 \pm 0.01$ & $0.024 \pm 0.000$ & $0.176 \pm 0.000$ & $0.943 \pm 0.002$ & $0.895 \pm 0.000$ \\
DNN + G-VBLL & $96.4 \pm 0.01$ & $0.035 \pm 0.000$ & $0.533 \pm 0.003$ & $0.729 \pm 0.004$ & $0.661 \pm 0.004$ \\
G-VBLL + MAP & $-$ & $-$ & $-$ & $0.950 \pm 0.006$ & $0.893 \pm 0.003$   \\     
\hline
Dropout  & $95.7 \pm 0.13$ & $0.013 \pm 0.002$ & $0.145 \pm 0.004$ & $0.934 \pm 0.004$ & $0.903 \pm 0.001$\\
Ensemble & $96.4 \pm 0.09$ & $0.011 \pm 0.092$ & $\bm{0.124 \pm 0.001}$ & $0.947 \pm 0.002$ & $\bm{0.914 \pm 0.000}$\\
BBB & $96.0 \pm 0.08$ & $0.033 \pm 0.001$  & $0.333 \pm 0.014$ & $0.957 \pm 0.004$ & $0.844 \pm 0.013$\\

\hline
DVI-IBLL (ours)& $\bm{96.9 \pm 0.12}$ & $0.018 \pm 0.001$ & $0.144 \pm 0.002$ & $\bm{0.972 \pm 0.006}$ & $0.906 \pm 0.005$\\
\hline
\end{tabular}
}
% \vspace{-.3cm}
\end{table*}

\begin{table*}[ht]
\caption{Results for Wide ResNet-28-10 on CIFAR-100.}\label{tab:cifar100results}
\centering
% \scriptsize
\resizebox{\linewidth}{!}{
\begin{tabular}{c|ccccc} 
Method & Accuracy ($\uparrow$) & ECE ($\downarrow$) & NLL ($\downarrow$) & SVHN AUC ($\uparrow$) & CIFAR-10 AUC ($\uparrow$)              \\ 
\hline
DNN & $80.4 \pm 0.29$  &  $0.107 \pm 0.004$ & $0.941 \pm 0.016$ & $0.799 \pm 0.020$ & $0.795 \pm 0.001$ \\ 
% \hline
SNGP  & $80.3 \pm 0.23$  & $\bm{0.030 \pm 0.004}$ & $0.761 \pm 0.007$ & $0.846 \pm 0.019$ & ${0.798 \pm 0.001}$ \\ 
% \hline
D-VBLL & $80.7  \pm 0.03$ & $0.040 \pm 0.002$ &  $0.913 \pm 0.011$  & $0.849 \pm 0.006$ & $0.791 \pm 0.003$ \\
G-VBLL & $80.4 \pm 0.10$  & $0.051 \pm 0.003$ & $0.945 \pm 0.009$  & $0.767 \pm 0.055$ & $0.752 \pm 0.015$ \\     
\hline
DNN + LL Laplace & $80.4 \pm 0.29$ & $0.210 \pm 0.018$ & $1.048 \pm 0.014$ & $0.834 \pm 0.014$ & $0.811 \pm 0.002$\\
DNN + D-VBLL & $80.7 \pm 0.02$ & $0.063 \pm 0.000$ & $0.831 \pm 0.005$ & $0.843 \pm 0.001$ & $0.804 \pm 0.001$\\
DNN + G-VBLL & $80.6 \pm 0.02$ & $0.186 \pm 0.003$ & $3.026 \pm 0.155$ & $0.638 \pm 0.021$ & $0.652 \pm 0.025$ \\
G-VBLL + MAP & $-$  & $-$ & $-$ & $0.793 \pm 0.032$ & $0.765 \pm 0.008$   \\   
\hline
Dropout  & $80.2 \pm 0.22$ & $0.031 \pm 0.002$ & $0.762 \pm 0.008$ & $0.800 \pm 0.014$ & $0.797 \pm 0.002$\\
Ensemble  & $\bm{82.5 \pm 0.19}$ & $0.041 \pm 0.002$ & $\bm{0.674 \pm 0.004}$ & $0.812 \pm 0.007$ & $\bm{0.814 \pm 0.001}$\\
BBB  &  $79.6 \pm 0.04$ & $0.127 \pm 0.002$  & $1.611 \pm 0.006$ & $0.809 \pm 0.060$ & $0.777 \pm 0.008$\\

\hline
DVI-IBLL (ours)  & $\bm{81.6 \pm 0.05}$ & $0.035\pm 0.002$ & $0.732 \pm 0.014$ & $\bm{0.854 \pm 0.006}$ & $0.804 \pm 0.004$\\
\hline
\end{tabular}
}
% \vspace{0.5cm}
\end{table*}
\section{Experiments}
\subsection{Metrics and Baselines}
In our regression experiments, we present the predictive negative log likelihood (NLL) for test data, which can be straightforwardly calculated for point feature estimates. Additionally, we include the root mean squared error (RMSE), a widely used metric in regression analysis. For classification tasks, apart from the negative log likelihood, we also showcase the predictive accuracy (based on the standard argmax of the predictive distribution) and the expected calibration error (ECE), which assesses the alignment between the model's subjective predictive uncertainty and actual predictive error. Furthermore, we delve into evaluating out-of-distribution detection performance, a recognized assessment approach for robust and probabilistic machine learning \cite{liu2023simple}. Specifically, we measure the area under the ROC curve (AUC) for datasets near and far from the distribution, which will be elaborated on further in this section.

In the realm of regression, we contrast our model with approaches that leverage exact conjugacy, including Bayesian last layer models such as GBLL and LDGBLL \cite{watson2021latent}, as well as  VBLL \cite{harrison2023variational} and RBF kernel Gaussian processes. We also juxtapose our model with MAP learning \cite{snoek2015scalable}, which involves training a complete network using MAP estimation and subsequently fitting a Bayesian last layer to these fixed features. 

In the domain of classification, our primary point of comparison is standard deep neural networks (DNN), given their ability to output a distribution over labels for direct comparison with our model. Additionally, we compare our model with SNGP \cite{liu2023simple} and last layer Laplace-based methods \cite{daxberger2021laplace}, which are akin to last layer models aiming to approximate deep kernel GPs \cite{wilson2016deep} and computing an approximate posterior after training, respectively. Note that Laplace methods are not assessed in regression due to their similarity to the MAP model. 

Furthermore, we examine various variational methods such as Bayes-by-Backprop (BBB) \cite{blundell2015weight}, Ensembles \cite{lakshminarayanan2017simple}, Bayesian Dropout \cite{gal2016dropout}, and Stochastic Weight Averaging-Gaussian (SWAG) \cite{maddox2019simple} for a comprehensive evaluation of performance. All our experiments were conducted on an RTX 4090 GPU.

\subsection{Regression}

Our UCI experiments closely align with the methods outlined in \cite{watson2021latent, harrison2023variational}, enabling a direct comparison with their baseline models. Consistently, we employed the same MLP architecture as described in \cite{watson2021latent}, comprising two layers of 50 hidden units each. Maintaining consistency with \cite{watson2021latent}, a batch size of 32 was utilized for all datasets, except for the Power dataset where a batch size of 256 was chosen to expedite training. Standard preprocessing techniques were applied, including normalization of inputs and subtraction of training set means from the outputs. The reported results in our manuscript exhibit the outcomes under leaky ReLU activations, with optimization performed using the AdamW optimizer \cite{loshchilov2017decoupled} across all experiments. 

In deterministic feature experiments, we conducted 20 runs with varying seeds. Each run involved splitting the data into training, validation, and testing sets with ratios of 0.72, 0.18, and 0.1, respectively. Training was executed on the training set, while performance monitoring on the validation set was carried out to determine the optimal number of epochs for model convergence. Our study delves into the performance analysis of regression VBLL models across 6 UCI regression datasets, as depicted in Tables \ref{tab:reg1results} and \ref{tab:reg2results}. Notably, our experiments unveil promising outcomes for DVI-IBLL models across diverse datasets, with the Gaussian process (GP) model showcasing competitive performance across the board. The results obtained from our UCI experiments underscore the substantial advancements achieved by our methodology compared to the baselines presented in \cite{watson2021latent, harrison2023variational}, thereby highlighting the enhanced flexibility of our BLL models in a compelling manner.
\subsection{Image Classification}

To evaluate the performance of our model in image classification on the CIFAR-10 and CIFAR-100 datasets using a Wide ResNet-28-10 backbone architecture, we investigated different training methods, post-training strategies, and feature uncertainty considerations. Our evaluation included assessing out-of-distribution (OOD) detection performance using the Street View House Numbers (SVHN) dataset as a far-OOD dataset and CIFAR-100 as a near-OOD dataset for CIFAR-10.

Our models exhibited strong  performance  in terms of Accuracy, SVHN AUC, and competitive metrics in terms of ECE, NLL and  SVHN metrics. These results showcase the excellent performance of our approach in image classification tasks. Additionally, we conducted an analysis of empirical runtime, with results presented in Table \ref{tab:runtimes}. The results indicate that our method introduced only a small increase in complexity compared to the baseline.

\begin{table}[b]
\caption{Time per batch on CIFAR-10 training.}

\centering
\begin{tabular}{c|cc} 

 Model & Run time (s) & \% above DNN \\ 

\hline

DNN & 0.321 & 0\% \\
D-VBLL & 0.338 & 5.2\%\\
G-VBLL & 0.364 & 13.4\%\\
DVI-IBLL (ours) & 0.343 & 6.9\%\\
\hline

\end{tabular}

\label{tab:runtimes}
\end{table}
\section{Conclusion}
In summary, our novel approach combining diffusion techniques and implicit priors for variational learning of Bayesian last layer weights enhances the expressive capacity of Bayesian Last Layer (BLL) models. This advancement addresses limitations with Gaussian priors and boosts model accuracy, calibration, and out-of-distribution detection proficiency in scenarios with non-Gaussian distributions, outliers, or high-dimensional data. Our method demonstrates potential for improving predictive accuracy and uncertainty quantification in challenging settings, promising enhanced performance of BLL models in machine learning tasks.

\newpage
\appendix
\section{Reproducibility Checklist}
This paper:

Includes a conceptual outline and/or pseudocode description of AI methods introduced (yes)

Clearly delineates statements that are opinions, hypothesis, and speculation from objective facts and results (yes)

Provides well marked pedagogical references for less-familiare readers to gain background necessary to replicate the paper (yes)

Does this paper make theoretical contributions? (yes)

If yes, please complete the list below.

All assumptions and restrictions are stated clearly and formally. (yes)

All novel claims are stated formally (e.g., in theorem statements). (yes)

Proofs of all novel claims are included. (yes)

Proof sketches or intuitions are given for complex and/or novel results. (yes)

Appropriate citations to theoretical tools used are given. (yes)

All theoretical claims are demonstrated empirically to hold. (yes)

All experimental code used to eliminate or disprove claims is included. (yes)

Does this paper rely on one or more datasets? (yes)

If yes, please complete the list below.

A motivation is given for why the experiments are conducted on the selected datasets (yes)

All novel datasets introduced in this paper are included in a data appendix. (NA)

All novel datasets introduced in this paper will be made publicly available upon publication of the paper with a license that allows free usage for research purposes. (NA)

All datasets drawn from the existing literature (potentially including authors’ own previously published work) are accompanied by appropriate citations. (yes)

All datasets drawn from the existing literature (potentially including authors’ own previously published work) are publicly available. (yes)

All datasets that are not publicly available are described in detail, with explanation why publicly available alternatives are not scientifically satisficing. (NA)

Does this paper include computational experiments? (yes)

If yes, please complete the list below.

Any code required for pre-processing data is included in the appendix. (yes).

All source code required for conducting and analyzing the experiments is included in a code appendix. (yes)

All source code required for conducting and analyzing the experiments will be made publicly available upon publication of the paper with a license that allows free usage for research purposes. (yes)

All source code implementing new methods have comments detailing the implementation, with references to the paper where each step comes from (yes)

If an algorithm depends on randomness, then the method used for setting seeds is described in a way sufficient to allow replication of results. (yes)

This paper specifies the computing infrastructure used for running experiments (hardware and software), including GPU/CPU models; amount of memory; operating system; names and versions of relevant software libraries and frameworks. (yes)

This paper formally describes evaluation metrics used and explains the motivation for choosing these metrics. (yes)

This paper states the number of algorithm runs used to compute each reported result. (yes)

Analysis of experiments goes beyond single-dimensional summaries of performance (e.g., average; median) to include measures of variation, confidence, or other distributional information. (yes)

The significance of any improvement or decrease in performance is judged using appropriate statistical tests (e.g., Wilcoxon signed-rank). (yes)

This paper lists all final (hyper-)parameters used for each model/algorithm in the paper’s experiments. (yes)

This paper states the number and range of values tried per (hyper-) parameter during development of the paper, along with the criterion used for selecting the final parameter setting. (yes)

\end{document}